\documentclass{article}
\usepackage{main, times}

\usepackage{amsmath,amsfonts,bm}

\def\eqref#1{equation~\ref{#1}}

\def\1{\bm{1}}

\DeclareMathAlphabet{\mathsfit}{\encodingdefault}{\sfdefault}{m}{sl}
\SetMathAlphabet{\mathsfit}{bold}{\encodingdefault}{\sfdefault}{bx}{n}

\usepackage{hyperref}
\usepackage{url}

\usepackage{multirow}
\usepackage{amsmath}
\usepackage{graphicx}
\usepackage{enumitem}
\usepackage{array}
\usepackage{booktabs}
\usepackage{caption}
\usepackage{subcaption}
\setlist[itemize]{leftmargin=*}

\title{Rethinking Unsupervised Cross-modal Flow Estimation: Learning from Decoupled Optimization and Consistency Constraint}

\author{
	Runmin Zhang$^{1}$\quad
	Jialiang Wang$^{1}$\quad
	Si-Yuan Cao$^{2,3}$\thanks{Corresponding author.}\quad
	Zhu Yu$^{1}$\quad \\[1pt]
	\textbf{Junchen Yu}$^{1}$\quad
	\textbf{Guangyi Zhang}$^{1}$\quad
	\textbf{Hui-Liang Shen}$^{1}$ \\[1pt]
	{\small{$^{1}$College of Information Science and Electronic Engineering, Zhejiang University}} \\[1pt]
	{\small{$^{2}$Ningbo Global Innovation Center, Zhejiang University}\quad {$^{3}$NingboTech University}} \\[1pt]
	{\tt \small{\{runmin\_zhang, cao\_siyuan\}@zju.edu.cn}}
}

\iclrfinalcopy
\begin{document}

\maketitle
\begin{abstract}
This work presents DCFlow, a novel unsupervised cross-modal flow estimation framework that integrates a decoupled optimization strategy and a cross-modal consistency constraint. Unlike previous approaches that implicitly learn flow estimation solely from appearance similarity, we introduce a decoupled optimization strategy with task-specific supervision to address modality discrepancy and geometric misalignment distinctly. This is achieved by collaboratively training a modality transfer network and a flow estimation network. To enable reliable motion supervision without ground-truth flow, we propose a geometry-aware data synthesis pipeline combined with an outlier-robust loss. Additionally, we introduce a cross-modal consistency constraint to jointly optimize both networks, significantly improving flow prediction accuracy. For evaluation, we construct a comprehensive cross-modal flow benchmark by repurposing public datasets. Experimental results demonstrate that DCFlow can be integrated with various flow estimation networks and achieves state-of-the-art performance among unsupervised approaches.
\end{abstract}

\section{Introduction}
Cross-modal flow estimation aims to establish pixel-wise correspondences between images captured from different modalities. It is crucial for various vision tasks, including multi-modal image fusion~\cite{dcevo_cvpr25}, image restoration~\cite{sgdformer_if25}, and depth estimation~\cite{vtd_tits23}. Due to the difficulty of acquiring ground-truth flow in real-world scenarios, unsupervised cross-modal flow estimation has attracted increasing attention.

Existing unsupervised approaches typically address this task by minimizing appearance discrepancies between image pairs. To achieve this, a modality transfer network is usually employed to translate images from one modality to another. For instance, NeMAR~\cite{nemar_cvpr20} simultaneously optimizes the modality transfer and flow estimation networks, while others~\cite{umf_ijcai22, rfnet_cvpr22} adopt a two-stage pipeline for separate training. Despite their different strategies, these methods share a fundamental limitation in implicitly learning flow estimation solely through appearance alignment. Consequently, they struggle particularly in textureless regions or repetitive structures, and their performance substantially degrades under large viewpoint changes due to the lack of direct flow supervision. This naturally raises a question: can we introduce reliable flow supervision using only unaligned cross-modal image pairs? Recent studies~\cite{lsfs_eccv20, mpiflow_iccv23} have explored generating synthetic motion labels from single images, showing promising results in mono-modal settings. However, whether such mono-modal supervision can benefit cross-modal scenarios, and how to effectively exploit it, remain largely unexplored.

\begin{figure}[t]
	\centering
	\includegraphics[width=0.95\columnwidth]{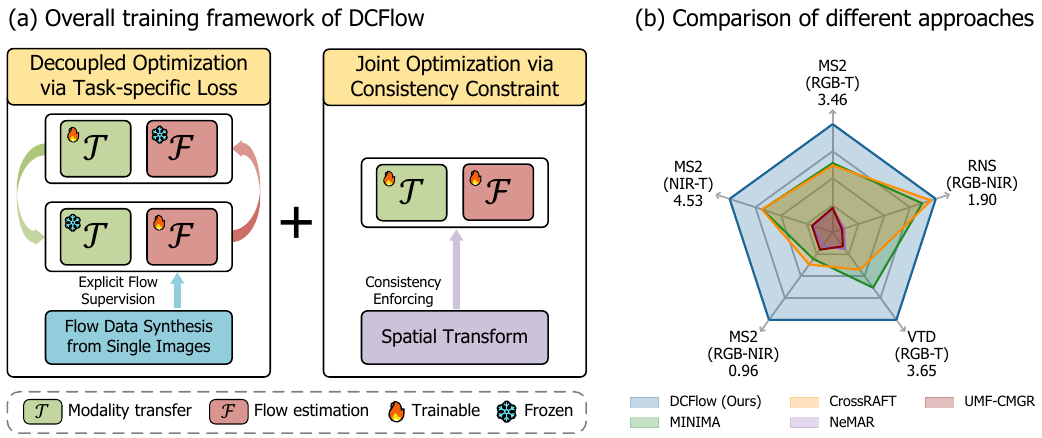}
	\caption{(a) Overall training framework of DCFlow, which integrates a decoupled optimization strategy and a cross-modal consistency constraint. (b) Comparison of the cross-modal flow estimation accuracy on five datasets. EPEs (endpoint errors) of different approaches are reported. Our DCFlow achieves state-of-the-art performance.}
	\label{fig:head}
\end{figure}

To bridge this gap, in this paper, we propose DCFlow, a novel framework for unsupervised cross-modal flow estimation that introduces effective flow supervision into the training process. In line with existing approaches, our framework consists of a modality transfer network and a flow estimation network. As shown in Fig.~\ref{fig:head}(a), DCFlow integrates a decoupled optimization strategy with task-specific supervision to train the two networks separately, as well as a cross-modal consistency constraint to jointly optimize them. Within this framework, we incorporate diverse flow supervision sources, including data synthesis from single images and spatial transformations on cross-modal image pairs.

The decoupled optimization strategy separates the overall task into modality transfer and single-modal flow estimation, forming a collaborative process where each network facilitates the optimization of the other. More importantly, it allows the flow network to be trained using only mono-modal flow supervision, while still contributing to accurate cross-modal alignment. To enable such supervision, we introduce a geometry-aware data synthesis pipeline that generates dense flow labels from single-view images. Considering that the synthetic data inevitably contains noise, we further design an outlier-robust loss to adaptively filter unreliable supervision based on residual magnitudes. These innovations enable effective flow network training without real-world labels, facilitating the decoupled training scheme. Compared to the conventional appearance-based optimization, our decoupled optimization strategy reduces the endpoint error (EPE) by 72\% on the MS$^2$ (RGB-T) dataset.

Furthermore, we propose a cross-modal consistency constraint to jointly optimize both the modality transfer and flow estimation networks. Specifically, we apply spatial transformations to cross-modal image pairs, and enforce consistency between flow predictions before and after transformations. This constraint encourages direct learning of cross-modal flow, and strengthens the mutual promotion of the two networks, resulting in a 28\% improvement in EPE on the MS$^2$ (RGB-T) dataset.

By integrating the above insights, DCFlow supports effective training of modern flow estimation networks such as RAFT~\cite{raft_eccv20} and FlowFormer~\cite{flowformer_eccv22}, offering a general and effective network-agnostic training framework for unsupervised cross-modal flow estimation. For comprehensive evaluation, we repurpose public multi-modal datasets by projecting LiDAR points to obtain ground-truth flow, creating five diverse datasets covering RGB, near-infrared (NIR), and thermal modalities. As shown in Fig.~\ref{fig:head}(b), DCFlow significantly surpasses existing unsupervised and large-scale pretrained approaches. In summary, our main contributions are as follows:

\begin{itemize}
	\item We propose DCFlow, a general and network-agnostic training framework for unsupervised cross-modal flow estimation. DCFlow achieves state-of-the-art performance among all unsupervised approaches.
	\item We introduce a decoupled optimization strategy that enables single-modal flow supervision to benefit cross-modal flow estimation, supported by a geometry-aware data synthesis pipeline and an outlier-robust loss to reliably provide such supervision from single-view images.
	\item We devise a cross-modal consistency constraint to facilitate effective joint optimization of the modality transfer and flow estimation networks, significantly enhancing flow estimation accuracy.
	\item We construct a comprehensive cross-modal flow benchmark by repurposing publicly available datasets, covering diverse modalities such as RGB, NIR, and thermal.
\end{itemize}

\section{Related Work}
\textbf{Cross-modal image matching.} Cross-modal image matching aims to establish spatial correspondences between images from different modalities, and has a wide range of applications~\cite{review_if21, fttp_arxiv23, promptfusion_as24, tris_aaai24}. Traditional approaches~\cite{rsncc_eccv14, dasc_cvpr15} design cross-modal invariant descriptors, but often suffer from high computational cost. Recently, unsupervised deep learning methods~\cite{nemar_cvpr20, umf_ijcai22, rfnet_cvpr22} have been proposed, typically consisting of a modality transfer network and a flow estimation network. Due to the absence of ground-truth correspondences, these methods rely on appearance-based supervision, which often leads to ambiguity in textureless or repetitive regions. Alternatively, approaches such as CrossRAFT~\cite{crossraft_aaai22} and MINIMA~\cite{minima_cvpr25} synthesize cross-modal data using multi-view RGB images with known ground-truth displacements to provide direct motion supervision. However, the synthetic-to-real domain gap limits their generalization to real-world scenarios. Besides, we note that recent work SSHNet~\cite{sshnet_cvpr25} introduces a split optimization framework for cross-modal homography estimation, which is limited to modeling global transformations. In contrast, we tackle the more challenging problem of dense pixel-wise correspondence estimation across modalities, which has broader applicability in real-world scenarios.

\textbf{Flow estimation.} Starting from FlowNet~\cite{flownet_iccv15}, various network architectures~\cite{raft_eccv20, gma_iccv21, flowformer_eccv22} have been proposed under supervised learning, with most state-of-the-art methods adopting an iterative prediction paradigm based on cost volumes. In the unsupervised setting, prior work~\cite{back_eccv16} introduces brightness constancy and motion smoothness as fundamental constraints. Subsequent approaches further enhance these ideas through specifically designed regularization strategies~\cite{lba_cvpr20, upflow_cvpr21, unsamflow_cvpr24}, or design domain adaption techniques to enhance performance in adverse weather~\cite{uda_aaai23, uda_cvpr23}. Other methods~\cite{lsfs_eccv20, mpiflow_iccv23, flowanthing_tpami25} attempt to synthesize training data from single-view images for large-scale training. However, most of the above approaches are designed for and evaluated on RGB image pairs, highlighting the urgent need for effective solutions tailored to real-world cross-modal scenarios.

\section{Method}
\subsection{Preliminaries}
This work tackles the problem of cross-modal flow estimation in an unsupervised setting. Given a cross-modal image pair $\mathbf{I}_\mathrm{A}$ and $\mathbf{I}_\mathrm{B}$ from modalities $\mathrm{A}$ and $\mathrm{B}$ respectively, our goal is to train a network $\mathcal{N}(\cdot)$ to predict the dense flow $\mathbf{F}_\mathrm{B2A}$ from $\mathbf{I}_\mathrm{B}$ to $\mathbf{I}_\mathrm{A}$. $\mathcal{N}(\cdot)$ is typically decomposed into two components, formulated as
\begin{align}
	\mathbf{F}_\mathrm{B2A} = \mathcal{N}(\mathbf{I}_\mathrm{A}, \mathbf{I}_\mathrm{B}) = \mathcal{F}_{\theta} (\mathcal{T}_{\phi}(\mathbf{I}_\mathrm{A}), \mathbf{I}_\mathrm{B}),
\end{align}
where $\mathcal{T}_\phi(\cdot)$ is a modality transfer network with learnable parameters $\phi$, which transforms $\mathbf{I}_\mathrm{A}$ into modality $\mathrm{B}$, and $\mathcal{F}_\theta(\cdot)$ is a mono-modal flow estimation network with learnable parameters $\theta$. Existing unsupervised approaches generally rely on photometric losses between the warped source and the target image for training, expressed as
\begin{align}
	\operatorname*{argmin}_{\phi, \theta} \mathcal{L}_\mathrm{ph} \left( \mathcal{W}(\mathbf{I}_\mathrm{A,T}, \mathbf{F}_\mathrm{B2A}), \mathbf{I}_\mathrm{B} \right),
	\label{eq:loss_ph}
\end{align}
where $\mathbf{I}_\mathrm{A,T}=\mathcal{T}_\phi(\mathbf{I}_\mathrm{A})$ is the modality transferred image, $\mathcal{W}(\cdot)$ denotes the warping operation, and $\mathcal{L}_\mathrm{ph}$ is a photometric similarity metric such as $\mathrm{L}_\mathrm{1}$ distance or SSIM. However, such supervision is inherently ambiguous in texture-less regions or repetitive patterns, and lacks direct motion cues, often leading to unsatisfactory results.

\begin{figure}[t]
	\centering
	\includegraphics[width=0.98\columnwidth]{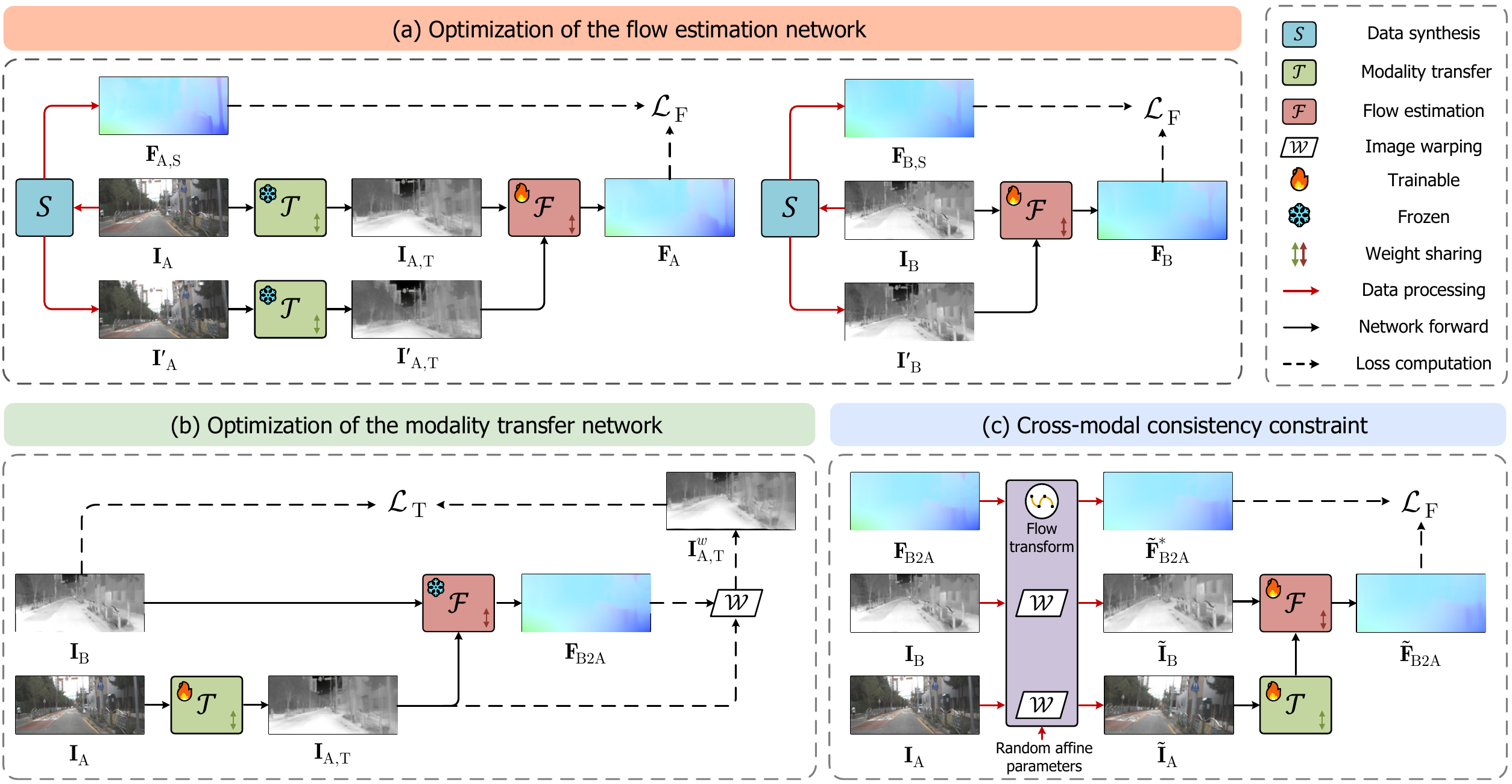}
	\caption{Schematic diagram of DCFlow, which incorporates a decoupled optimization strategy (a, b) and a cross-modal consistency constraint (c). (a) Optimization of the flow estimation network $\mathcal{F}_\theta$, where $\mathcal{F}_\theta$ is optimized using direct flow supervision from two-branch intra-modal synthetic data. (b) Optimization of the modality transfer network $\mathcal{T}_\phi$, where $\mathcal{T}_\phi$ learns to align modality $\mathbf{I}_\mathrm{A}$ with $\mathbf{I}_\mathrm{B}$ via perceptual similarity. (c) Cross-modal consistency constraint, where both networks are jointly optimized by enforcing flow consistency under known spatial transformations.}
	\label{fig:method}
\end{figure}

To overcome these limitations, we propose DCFlow, a novel unsupervised training framework that provides explicit flow supervision without requiring any labeled cross-modal data. As illustrated in Fig.~\ref{fig:method}, DCFlow consists of a decoupled optimization strategy that separately trains the modality transfer and flow estimation networks (Sec.~\ref{subsec:decoupled_optim}), and a cross-modal consistency constraint for joint optimization (Sec.~\ref{subsec:consistency_constraint}). To enable reliable flow supervision, we further introduce a geometry-aware data synthesis pipeline coupled with an outlier-robust loss (Sec.~\ref{subsec:flow_data}).

\subsection{Decoupled Optimization}
\label{subsec:decoupled_optim}
DCFlow adopts a decoupled optimization strategy for the modality transfer network $\mathcal{T}_\phi$ and the flow estimation network $\mathcal{F}_\theta$, which address the modality discrepancy and geometric misalignment between the input image pair separately, as illustrated in Fig.~\ref{fig:method}(a)(b). Each network is trained with task-specific supervision, forming a self-reinforcing process where improvements in one network provide better guidance for the other. Consequently, the training process becomes more stable, and converges to more accurate results.

\textbf{Flow estimation.} Fig.~\ref{fig:method}(a) illustrates the optimization process of the flow estimation network. In this stage, the weights of the modality transfer network $\mathcal{T}_\phi$ are frozen. The objective is to provide direct motion supervision for the flow network $\mathcal{F}_\theta$, addressing the limitations of implicit appearance-based supervision. Inspired by recent works~\cite{scpnet_eccv24, sshnet_cvpr25}, we adopt a two-branch intra-modal supervision scheme using synthetic data from each modality. It enables the flow estimation network to process inputs from two different domains simultaneously at the start of training. Under the multi-task learning paradigm~\cite{multitask_ml97}, this setup progressively enhances the generalization of the network to cross-modal inputs, and facilitates convergence in the unsupervised process.

Specifically, we construct two training triplets $(\mathbf{I}_\mathrm{A}, \mathbf{I}'_\mathrm{A}, \mathbf{F}_\mathrm{A,S})$ and $(\mathbf{I}_\mathrm{B}, \mathbf{I}'_\mathrm{B}, \mathbf{F}_\mathrm{B,S})$, where $\mathbf{I}'_\mathrm{A/B}$ denotes the rendered novel view from $\mathbf{I}_\mathrm{A/B}$, and $\mathbf{F}_\mathrm{A/B,S}$ is the corresponding synthetic flow label. The data synthesis pipeline is detailed in Sec.~\ref{subsec:flow_data}. The flow network is trained by minimizing the $\mathrm{L}_\mathrm{1}$ distance between the predicted and synthetic flows from both branches, formulated as
\begin{align}
	\operatorname*{argmin}_{\theta} (\mathcal{L}_\mathrm{F}(\mathbf{F}_\mathrm{A}, \mathbf{F}_\mathrm{A,S}) + \mathcal{L}_\mathrm{F}(\mathbf{F}_\mathrm{B}, \mathbf{F}_\mathrm{B,S})),
\end{align}
where the predicted flows are defined as $\mathbf{F}_\mathrm{A} = \mathcal{F}_{\theta}(\mathcal{T}_{\phi}(\mathbf{I}'_\mathrm{A}), \mathcal{T}_{\phi}(\mathbf{I}_\mathrm{A}))$ and $\mathbf{F}_\mathrm{B} = \mathcal{F}_{\theta}(\mathbf{I}'_\mathrm{B}, \mathbf{I}_\mathrm{B})$, and the loss $\mathcal{L}_\mathrm{F}$ is computed as
\begin{align}
	\mathcal{L}_\mathrm{F}(\mathbf{F}, \mathbf{F}_\mathrm{S}) = \left\| \mathbf{F} - \mathbf{F}_\mathrm{S} \right\|_1.
\end{align}
We implement $\mathcal{F}_\theta$ using RAFT~\cite{raft_eccv20}, an iterative flow estimation architecture, and apply the loss over all intermediate predictions. Notably, DCFlow is agnostic to the choice of flow network, and supports alternative architectures, as demonstrated in Table~\ref{table:ablation_flownet}.

\textbf{Modality transfer.} Fig.~\ref{fig:method}(b) illustrates the optimization process of the modality transfer network. In this stage, the weights of the flow estimation network $\mathcal{F}_\theta$ are frozen. The modality transfer network $\mathcal{T}_\phi$ is optimized to translate the input image $\mathbf{I}_\mathrm{A}$ into the appearance of modality $\mathrm{B}$.

Specifically, an estimated flow $\mathbf{F}_\mathrm{B2A}$ between the cross-modal image pair is used to warp the transferred image $\mathbf{I}_\mathrm{A,T}$, producing the warped output $\mathbf{I}_\mathrm{A,T}^w = \mathcal{W}(\mathbf{I}_\mathrm{A,T}, \mathbf{F}_\mathrm{B2A})$. The optimization objective is then given by
\begin{align}
	\operatorname*{argmin}_{\phi} \mathcal{L}_\mathrm{T}(\mathbf{I}_\mathrm{A,T}^w, \mathbf{I}_\mathrm{B}),
\end{align}
where $\mathcal{L}_\mathrm{T}$ is defined as the perceptual loss~\cite{perceptual_eccv16}, formulated as
\begin{align}
	\mathcal{L}_\mathrm{T}(\mathbf{I}_\mathrm{A,T}^w, \mathbf{I}_\mathrm{B}) = \sum_{l} \lambda_l \left\| \Phi_l(\mathbf{I}_\mathrm{A,T}^w) - \Phi_l(\mathbf{I}_\mathrm{B}) \right\|_2,
\end{align}
with $\Phi_l(\cdot)$ denoting the $l$-th layer of a pretrained VGG network, and $\lambda_l$ the corresponding layer weight. The perceptual loss captures high-level structural and semantic similarity, and is less sensitive to spatial misalignments than pixel-wise metrics like $\mathrm{L}_1$ distance. This makes it a robust supervisory signal for the modality transfer network, even when the estimated flow is imperfect, thereby guiding the unsupervised training toward a desired convergence. We implement $\mathcal{T}_\phi$ using a U-Net~\cite{unet_miccai15} to preserve both fine-grained details and global contextual features.

Under this decoupled training strategy, the two networks are trained independently with stable supervision. Meanwhile, their outputs mutually reinforce each other, driving the entire framework toward continuous performance improvement. Better still, this strategy enables the flow network to be trained using only mono-modal supervision, while contributing to cross-modal alignment, fundamentally addressing limitations of appearance-based methods.

\subsection{Data Synthesis and Outlier-robust Loss}
\label{subsec:flow_data}
\begin{figure}[t]
	\centering
	\includegraphics[width=0.8\columnwidth]{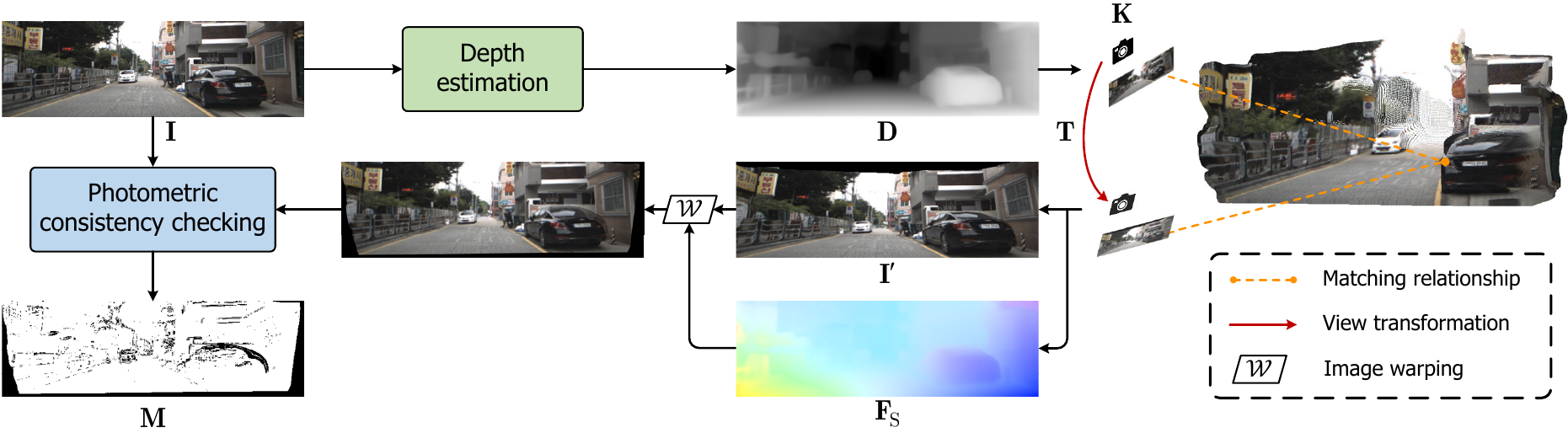}
	\caption{Illustration of the geometry-aware data synthesis pipeline.}
	\label{fig:dibr}
\end{figure}

To provide reliable supervision for the flow estimation network, we propose a geometry-aware data synthesis pipeline with an outlier-robust loss. The data synthesis process aims to generate a novel view image $\mathbf{I}'$ and its corresponding synthetic flow $\mathbf{F}_\mathrm{S}$ from a single input image $\mathbf{I}$. To achieve this, we introduce a lifting and reprojection technique that projects 2D pixels into 3D space and reprojects them into a virtual camera view. It produces photorealistic and geometrically consistent image pairs under realistic motion patterns, providing dense, geometry-grounded supervision without requiring multi-view images with ground-truth flow label. Fig.~\ref{fig:dibr} illustrates the overall pipeline.

We first estimate a depth map $\mathbf{D}$ using a pretrained monocular depth model, such as UniDepth~\cite{unidepth_cvpr24}. Each 2D pixel $\mathbf{x} = [u, v]^\top$ is projected into a 3D point $\mathbf{X} = [x, y, z]^\top$ via a sampled intrinsic matrix $\mathbf{K}$ as
\begin{align}
	\mathbf{X} \sim \mathbf{D}(\mathbf{x}) \cdot \mathbf{K}^{-1} \cdot \mathbf{x}.
\end{align}
For simplicity, we omit the homogeneous coordinate form. We then sample a virtual camera pose $\mathbf{T} \in \mathrm{SE}(3)$, and re-project the 3D points into the novel view as
\begin{align}
	\mathbf{x}' \sim \mathbf{K} \cdot \mathbf{T} \cdot \mathbf{X},
\end{align}
where $\mathbf{x}' = [u', v']^\top$. Then, the rendered image $\mathbf{I}'$ is obtained by sampling the corresponding pixel values from $\mathbf{I}$, and the synthetic flow $\mathbf{F}_\mathrm{S}$ is defined as the 2D displacement between $\mathbf{x}$ and $\mathbf{x}'$.

To identify occluded and invisible regions introduced by viewpoint changes, we adopt a photometric consistency check to generate a binary valid mask $\mathbf{M}$, where $\mathbf{M} = 1$ indicates valid pixels. Specifically, we warp the rendered image $\mathbf{I}'$ back to the original view using $\mathbf{F}_\mathrm{S}$, and measure the photometric error against $\mathbf{I}$. Pixels with photometric errors exceeding a threshold are marked as invalid and excluded from loss computation. The flow loss is then defined as
\begin{align}
	\mathcal{L}_\mathrm{F} = \frac{1}{\sum \mathbf{M}} \sum_{\mathbf{M}(\mathbf{x})=1} \left\| \mathbf{F}(\mathbf{x}) - \mathbf{F}_\mathrm{S}(\mathbf{x}) \right\|_1.
\end{align}
While the valid mask effectively excludes occluded and invisible regions, some pixels may still yield unreliable supervision due to rendering artifacts or depth estimation errors. These regions, such as distorted boundaries or thin structures, may contain valid flow but poor appearance quality, making them unsuitable for flow supervision.

To address this, we introduce an outlier-robust loss that further filters noisy supervision by discarding a small fraction of pixels with the highest residuals. Specifically, we sort the per-pixel $\mathrm{L}_1$ flow error within valid regions, and remove the top-$\tau\%$ pixels for loss computation. This ensures that the supervision focuses on regions with both accurate motion and relatively high appearance fidelity. The outlier-robust loss is defined as
\begin{align}
	\mathcal{L}_\mathrm{F} = \frac{1}{|\Omega_\tau|} \sum_{\mathbf{x} \in \Omega_\tau} \left\| \mathbf{F}(\mathbf{x}) - \mathbf{F}_\mathrm{S}(\mathbf{x}) \right\|_1,
	\label{eq:loss_or}
\end{align}
where $\Omega_\tau$ denotes the set of valid pixels after excluding the top-$\tau\%$ largest residuals. It provides a more robust training signal by avoiding noisy supervision caused by uncertain depth or texture distortions, thereby further improving the stability and accuracy of flow learning.

\subsection{Cross-modal Consistency Constraint}
\label{subsec:consistency_constraint}
Though the decoupled optimization strategy ensures stable convergence, it does not explicitly teach the network to perform cross-modal flow estimation. Instead, the model relies on the generalization ability of two independently optimized networks. To overcome this limitation, we introduce a cross-modal consistency constraint that jointly optimizes both networks by enforcing flow consistency under known spatial transformations, as illustrated in Fig.~\ref{fig:method}(c).

Given a cross-modal image pair $(\mathbf{I}_\mathrm{A}, \mathbf{I}_\mathrm{B})$ and the corresponding flow prediction $\mathbf{F}_\mathrm{B2A}$, we apply random affine transformations (\emph{e.g.}, scaling, rotation, and translation) to both images, yielding the augmented image pair $(\tilde{\mathbf{I}}_\mathrm{A}, \tilde{\mathbf{I}}_\mathrm{B})$. Since the transformations are known, we can derive a transformed flow $\tilde{\mathbf{F}}_\mathrm{B2A,T}^{*}$ from $\mathbf{F}_\mathrm{B2A}$. The augmented pair is then passed through the full network pipeline, producing a new prediction $\tilde{\mathbf{F}}_\mathrm{B2A} = \mathcal{F}_{\theta}(\mathcal{T}_\phi(\tilde{\mathbf{I}}_\mathrm{A}), \tilde{\mathbf{I}}_\mathrm{B})$. Based on the assumption that flow predictions should remain geometrically consistent under known transformations, the joint optimization objective is defined as
\begin{align}
	\operatorname*{argmin}_{\phi, \theta} \mathcal{L}_\mathrm{F}(\tilde{\mathbf{F}}_\mathrm{B2A}, \tilde{\mathbf{F}}_\mathrm{B2A}^{*}),
\end{align}
where $\mathcal{L}_\mathrm{F}$ is the outlier-robust loss in Eq.~\ref{eq:loss_or}. This self-supervised constraint enables direct learning of cross-modal flow estimation, encouraging mutual adaptation between the two networks and thus improving final performance.

\subsection{Overall Training Objective}
\label{subsec:overall}
The entire training objective can be formulated as
\begin{equation}
	\begin{aligned}
		\operatorname*{argmin}_{\theta} & \ (\mathcal{L}_\mathrm{F}(\mathbf{F}_\mathrm{A}, \mathbf{F}_\mathrm{A,S}) + \mathcal{L}_\mathrm{F}(\mathbf{F}_\mathrm{B}, \mathbf{F}_\mathrm{B,S})) \\
		+\operatorname*{argmin}_{\phi} & \ \lambda_\mathrm{T} \mathcal{L}_\mathrm{T}(\mathbf{I}_\mathrm{A,T}^w, \mathbf{I}_\mathrm{B})
		+\operatorname*{argmin}_{\phi, \theta} \ \lambda_\mathrm{C} \mathcal{L}_\mathrm{F}(\tilde{\mathbf{F}}_\mathrm{B2A}, \tilde{\mathbf{F}}_\mathrm{B2A}^{*}),
	\end{aligned}
\end{equation}
where $\lambda_\mathrm{T}$ and $\lambda_\mathrm{C}$ denote the loss weights for the modality transfer and cross-modal consistency components, respectively. We note that these three losses are jointly optimized within a single gradient descent step, 
with each loss imposed on its corresponding set of parameters. Once combined, the decoupled optimization allows each network to learn its specific task robustly, while the consistency constraint enhances collaboration between two networks for better cross-modal flow estimation, ultimately leading to stable and effective unsupervised training.

For implementation, We set $\lambda_\mathrm{T} = 2.0$ and $\lambda_\mathrm{C} = 0.05$. We train the entire network from scratch for $30,000$ iterations with a batch size of $4$. The cross-modal consistency constraint is introduced after $10,000$ iterations, allowing the model to produce reliable flow predictions before joint optimization begins.

\begin{table}[t]
	\centering
	\caption{The statistics of cross-modal flow datasets.}
	\renewcommand{\arraystretch}{1.1}
	\resizebox{0.55\columnwidth}{!}{
		\begin{tabular}{c|cccc}
			\specialrule{1pt}{0ex}{0.2ex}
			Dataset & Modality & \# Training & \# Testing & Resolution \\
			\hline
			MS$^2$ & RGB-T-NIR & 3631 & 907 & 608$\times$192 \\
			VTD & RGB-T & 1536 & 392 & 512$\times$320 \\
			RNS & RGB-NIR & 1187 & 294 & 704$\times$512 \\
			\specialrule{1pt}{0.2ex}{0ex}
		\end{tabular}
	}
	\label{table:dataset}
\end{table}

\section{Experiments}
\subsection{Experimental Setting}
\textbf{Datasets.} We evaluate DCFlow on three public datasets, \emph{i.e.}, MS$^2$~\cite{ms2_cvpr23}, VTD~\cite{vtd_tits23}, and RNS~\cite{rns_cvpr25}, These datasets provide multi-modal data including RGB, near-infrared (NIR), thermal (T), and LiDAR, and consist of multiple video sequences. To adapt them for cross-modal flow evaluation, we repurpose the raw data using the following procedure. We first resize and crop images across modalities to achieve consistent effective focal lengths, based on the provided intrinsic parameters. Then, we project LiDAR points from one image to another using the known extrinsic parameters, and compute the ground-truth flow from the 2D displacements of valid projected points. Due to the inherent sparsity of LiDAR, the resulting flow labels are sparse. For dataset splitting, we use the first 80\% frames in each video sequence for training and the remaining 20\% for testing. Dataset statistics are summarized in Table~\ref{table:dataset}. Notably, the MS$^2$ dataset simultaneously captures RGB, NIR, and T modalities, allowing us to construct three sub-datasets for comprehensive evaluation under different modality gaps. We denote these as MS$^2$ (RGB-T), MS$^2$ (RGB-NIR), and MS$^2$ (NIR-T).

\textbf{Metrics.} We report the endpoint error (EPE) and the flow outlier rate (F1). EPE measures the average $\mathrm{L}_2$ distance between the predicted flow and the ground-truth, while F1 denotes the percentage of pixels with EPE greater than both 3 pixels and 5\% of the ground-truth magnitude. Lower EPE and F1 values indicate better performance.

\begin{figure}[t]
	\centering
	\includegraphics[width=1\columnwidth]{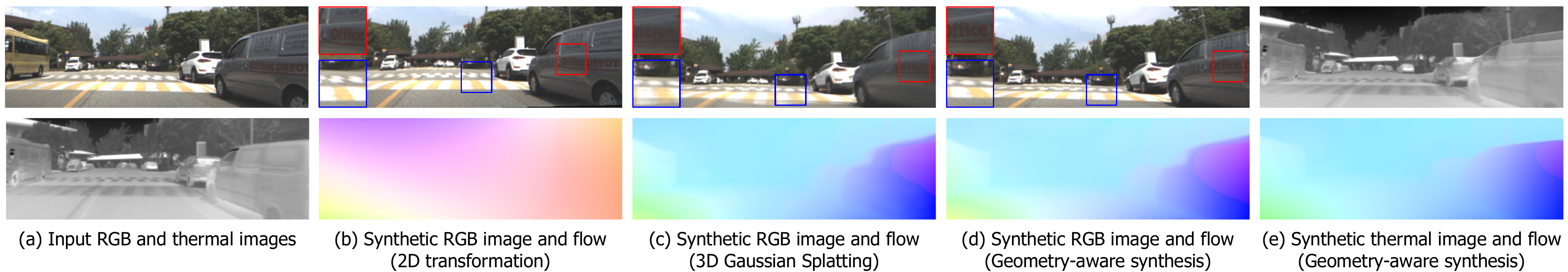}
	\caption{Qualitative comparison of different synthetic flow data generation strategies.}
	\label{fig:ablation_syn}
\end{figure}

\begin{table}[t]
	\centering
	\caption{Ablation studies of DCFlow on the MS$^2$ (RGB-T) dataset.}
	\begin{subtable}[t]{0.48\linewidth}
		\centering
		\caption{Ablation study on training strategies.}
		\renewcommand{\arraystretch}{1.1}
		\resizebox{\linewidth}{!}{
			\begin{tabular}{l|cc}
				\specialrule{1pt}{0ex}{0.2ex}
				Training strategy & EPE & F1 \\
				\hline
				Appearance-based optimization & 21.23 & 98.45 \\
				\hline
				Decoupled optimization & 5.80 & 57.18 \\
				+ Outlier-robust loss & 4.81 & 51.39 \\
				+ Cross-modal consistency constraint & \textbf{3.46} & \textbf{35.89} \\
				\specialrule{1pt}{0.2ex}{0ex}
			\end{tabular}
		}
		\label{table:ablation_train}
	\end{subtable}
	\vspace{2mm}
	\hfill
	\begin{subtable}[t]{0.48\linewidth}
		\centering
		\caption{Ablation study on data synthesis strategies.}
		\renewcommand{\arraystretch}{1.1}
		\resizebox{\linewidth}{!}{
			\begin{tabular}{c|cc}
				\specialrule{1pt}{0ex}{0.2ex}
				Data synthesis strategy & EPE & F1 \\
				\hline
				2D transformation & 13.12 & 95.21 \\
				3D Gaussian Splatting & 5.11 & 63.12 \\
				Geometry-aware synthesis (Ours) & \textbf{3.46} & \textbf{35.89} \\
				\specialrule{1pt}{0.2ex}{0ex}
			\end{tabular}
		}
		\label{table:ablation_syn}
	\end{subtable}
	\begin{subtable}[t]{0.55\linewidth}
		\centering
		\caption{Ablation study on different flow networks.}
		\renewcommand{\arraystretch}{1.1}
		\resizebox{\linewidth}{!}{
			\begin{tabular}{c|cccc}
				\specialrule{1pt}{0ex}{0.2ex}
				Flow network & RAFT & GMA & FlowFormer & SEA-RAFT \\
				\hline
				EPE & 3.46 & 4.13 & 3.66 & 3.57 \\
				F1  & 35.89 & 48.04 & 37.97 & 38.63 \\
				\specialrule{1pt}{0.2ex}{0ex}
			\end{tabular}
		}
		\label{table:ablation_flownet}
	\end{subtable}
\end{table}

\subsection{Ablation Study}
Ablation studies are conducted on the MS$^2$ (RGB-T) dataset.

\textbf{Training strategy.} Table~\ref{table:ablation_train} presents the ablation study on different training strategies of DCFlow. We start with the appearance-based optimization baseline, where the entire network is trained using only the photometric similarity loss in Eq.~\ref{eq:loss_ph}. As shown, this strategy leads to poor performance, reflecting the inherent limitations of relying solely on appearance cues. We then evaluate our decoupled optimization strategy, which separately trains the modality transfer and flow estimation networks using task-specific objectives. In this setup, synthetic flow data is introduced to enable explicit motion supervision for the flow network. This strategy yields stable convergence and achieves an EPE of 5.80. Introducing the outlier-robust loss further improves performance, reducing EPE by 0.99 and F1 by 5.79, demonstrating its effectiveness in suppressing noisy supervision from synthetic artifacts. Finally, incorporating the proposed cross-modal consistency constraint leads to the best performance, with an EPE of 3.46 and F1 of 35.89. These results confirm that enforcing spatial consistency during joint optimization significantly enhances flow estimation accuracy.

\textbf{Flow data synthesis.} DCFlow introduces a geometry-aware synthesis pipeline to generate image pairs with dense flow labels from a single image. As alternatives, we evaluate the 2D transformation (\emph{e.g.}, homography) and feed-forward 3D Gaussian Splatting~\cite{flash3d_3dv25} as substitute data generation strategies. We present qualitative comparisons of these strategies in Fig.~\ref{fig:ablation_syn}. The 2D transformation (Fig.~\ref{fig:ablation_syn}(b)) ignores scene geometry and produces unrealistic motion patterns, which cause the model to overfit to such distortions. The 3D Gaussian Splatting approach (Fig.~\ref{fig:ablation_syn}(c)) synthesizes novel views with 3D awareness, but often suffers from visual artifacts and instability caused by imperfect Gaussian primitives estimation. In contrast, our proposed data synthesis pipeline (Fig.~\ref{fig:ablation_syn}(d)) produces geometrically consistent and visually plausible novel views, offering more reliable training data for flow networks. Moreover, Fig.~\ref{fig:ablation_syn}(e) shows that our geometry-aware synthesis pipeline generalizes well to challenging modalities such as thermal images. Table~\ref{table:ablation_syn} reports the results trained under each data synthesis pipeline, further demonstrating that our geometry-aware synthesis significantly outperforms the alternatives.

\textbf{Generalization ability for different flow networks.} We replace RAFT~\cite{raft_eccv20} with GMA~\cite{gma_iccv21}, FlowFormer~\cite{flowformer_eccv22}, and SEA-RAFT~\cite{searaft_eccv24}, and report the results in Table~\ref{table:ablation_flownet}. Our DCFlow consistently achieves superior performance across different flow networks, demonstrating strong generalization and compatibility.

\subsection{Comparisons with Existing Approaches}

\textbf{Baselines.} We evaluate our DCFlow with large-scale pretrained approaches including CrossRAFT~\cite{crossraft_aaai22} and MINIMA~\cite{minima_cvpr25}, unsupervised approaches including NeMAR~\cite{nemar_cvpr20} and UMF-CMGR~\cite{umf_ijcai22}, and supervised approaches including RAFT~\cite{raft_eccv20}, GMA~\cite{gma_iccv21}, FlowFormer~\cite{flowformer_eccv22}, and SEA-RAFT~\cite{searaft_eccv24}. CrossRAFT and MINIMA are pretrained on large-scale synthetic datasets with ground-truth flow label. We evaluate their performance using publicly available checkpoints. We retrain all unsupervised and supervised baselines under the same settings as ours for fairness. The supervised approaches are trained using sparse ground-truth flow annotations.

\begin{figure*}[t]
	\centering
	\includegraphics[width=1\columnwidth]{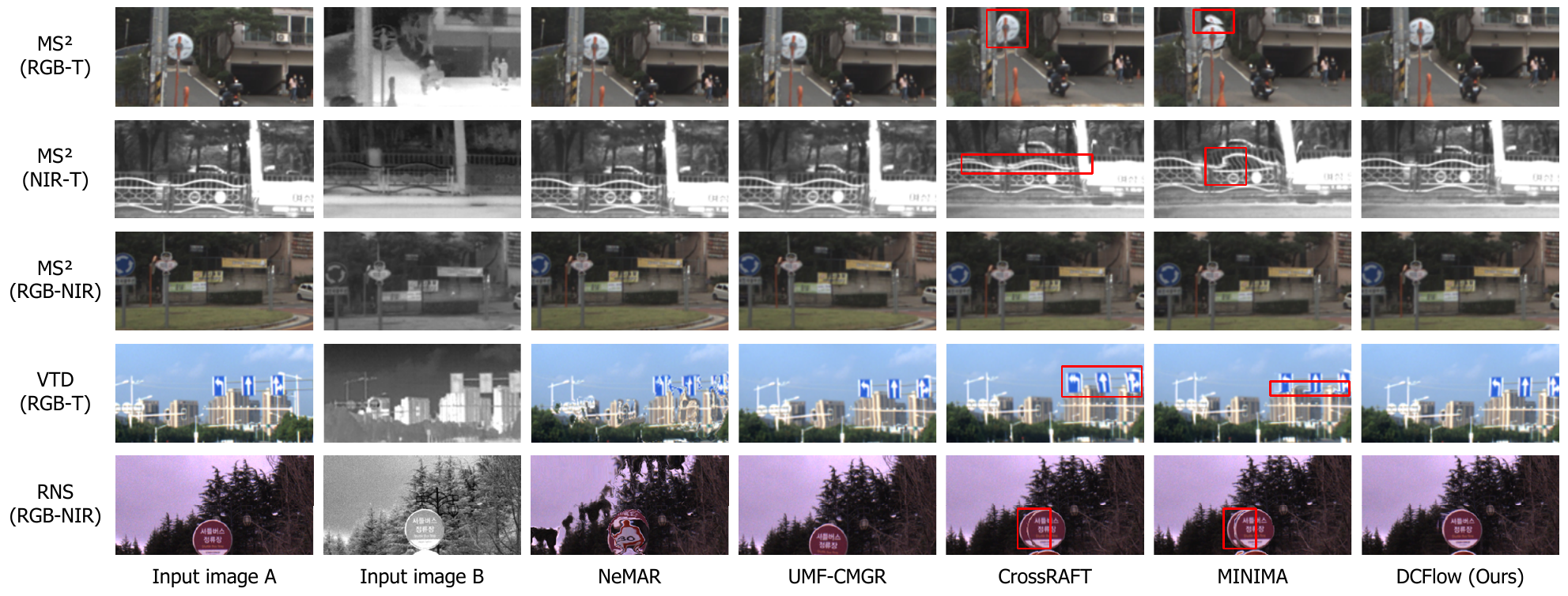}
	\vspace{-4mm}
	\caption{Qualitative comparison of DCFlow and other approaches. The first two columns show the input image pairs, and the remaining columns visualize the image from modality $\mathrm{A}$ warped using the estimated flow from each approach. For clarity, cropped patches of the full-resolution images are shown. The red boxes highlight the distortion regions.}
	\vspace{-2mm}
	\label{fig:visres}
\end{figure*}

\begin{table}[t]
	\centering
	\caption{Quantitative comparison of DCFlow and other approaches. The EPE and F1 are reported. The best results among all large-scale pretrained and unsupervised methods are highlighted in bold.}
	\vspace{-2mm}
	\resizebox{\linewidth}{!}
	{   
		\renewcommand{\arraystretch}{1.1}
		\begin{tabular}{c|c|cc|cc|cc|cc|cc}
			\specialrule{1pt}{0.4ex}{0.4ex}
			\multirow{3}{*}{Category} &\multirow{3}{*}{Method} & \multicolumn{2}{c|}{MS$^2$} & \multicolumn{2}{c|}{MS$^2$} & \multicolumn{2}{c|}{MS$^2$} & \multicolumn{2}{c|}{VTD} & \multicolumn{2}{c}{RNS} \\
			& & \multicolumn{2}{c|}{(RGB-T)} & \multicolumn{2}{c|}{(NIR-T)} & \multicolumn{2}{c|}{(RGB-NIR)} & \multicolumn{2}{c|}{(RGB-T)} & \multicolumn{2}{c}{(RGB-NIR)} \\  
			\cline{3-12}
			& & EPE & F1 & EPE & F1 & EPE & F1 & EPE & F1 & EPE & F1 \\
			\hline
			\multirow{4}{*}{Supervised} & RAFT & 1.70 & 14.76 & 1.80 & 16.50 & 0.23 & 0.57 & 1.14 & 8.45 & 1.63 & 9.32 \\
			& GMA & 1.67 & 14.58 & 1.80 & 16.55 & 0.24 & 0.61 & 1.15 & 8.67 & 1.53 & 6.53 \\
			& FlowFormer & 1.65 & 14.28 & 1.78 & 16.22 & 0.25 & 0.77 & 1.20 & 8.43 & 1.87 & 13.88 \\
			& SEA-RAFT & 1.65 & 14.04 & 1.97 & 17.64 & 0.21 & 0.48 & 1.31 & 9.49 & 1.37 & 6.42 \\
			\hline
			Large-scale & CrossRAFT & 6.21 & 70.20 & 7.06 & 73.06 & 4.32 & 29.54 & 9.86 & 89.33 & 2.04 & 15.53 \\
			pretrained & MINIMA & 5.97 & 66.12 & 7.10 & 70.37 & 5.44 & 33.48 & 6.34 & 80.41 & 2.34 & 15.18 \\
			\hline
			\multirow{3}{*}{Unsupervised} & NeMAR & 19.25 & 99.80 & 28.41 & 99.99 & 11.39 & 99.92 & 23.43 & 96.89 & 25.11 & 99.83 \\
			& UMF-CMGR & 18.84 & 99.78 & 26.67 & 99.99 & 8.85 & 99.27 & 28.05 & 99.13 & 31.13 & 99.98\\
			& DCFlow (Ours) & \textbf{3.46} & \textbf{35.89} & \textbf{4.53} & \textbf{49.81} & \textbf{0.96} & \textbf{5.00} & \textbf{3.65} & \textbf{48.49} & \textbf{1.90} & \textbf{13.05} \\
			\specialrule{1pt}{0.4ex}{0.4ex}
		\end{tabular}
	}
	\vspace{-4mm}
	\label{table:cmp}
\end{table}

\textbf{Quantitative comparison.}
Table~\ref{table:cmp} reports the quantitative results on five cross-modal datasets. Among all unsupervised and large-scale pretrained approaches, our DCFlow consistently achieves the best performance on both metrics, demonstrating strong generalization ability across diverse modalities. The existing unsupervised approaches such as NeMAR and UMF-CMGR generally yield unsatisfactory results, which aligns with findings from prior studies~\cite{scpnet_eccv24, sshnet_cvpr25}. This suggests that appearance-based optimization struggles to converge under significant modality discrepancy and geometric misalignment. In contrast, DCFlow produces stable and accurate flow estimation, highlighting the effectiveness of our training framework. Compared to large-scale pretrained approaches like CrossRAFT and MINIMA, DCFlow achieves significantly better results. For instance, DCFlow yields 42.0\%, 36.2\%, 82.4\%, 42.4\%, and 18.8\% lower EPEs than MINIMA on the five datasets, demonstrating the advantage of learning directly from unlabeled real-world data over relying on synthetic cross-modal datasets. When compared with supervised approaches, our DCFlow achieves competitive performance, despite the absence of using ground-truth cross-modal flow labels. These results highlight the effectiveness of DCFlow. 

\textbf{Qualitative comparison.} Fig.~\ref{fig:visres} presents qualitative results on five datasets. We visualize the image from modality $\mathrm{A}$ warped using the estimated flow to assess the accuracy of each method. As shown, previous unsupervised approaches such as NeMAR and UMF-CMGR struggle with large cross-modal misalignments. Although CrossRAFT and MINIMA produce coarse alignment, they still exhibit noticeable mismatches in several regions, as highlighted by the red boxes. In contrast, DCFlow achieves more precise warping, demonstrating the superiority of our training framework.

\section{Conclusions}
We have presented DCFlow, a novel unsupervised framework for cross-modal flow estimation that combines a decoupled optimization strategy and a cross-modal consistency constraint. The former tackles modality discrepancy and geometric misalignment with task-specific supervision, while the latter enables direct learning of cross-modal flow. Within this framework, we introduce a geometry-aware synthesis pipeline with an outlier-robust loss to provide reliable flow supervision from single-view images. Experiments on multiple datasets demonstrate that DCFlow achieves state-of-the-art performance among unsupervised approaches.

\bibliography{main}
\bibliographystyle{main}

\end{document}